\documentclass{article}
\usepackage{ijcai24}

\usepackage{times}
\usepackage{soul}
\usepackage{url}
\usepackage[hidelinks]{hyperref}
\usepackage[utf8]{inputenc}
\usepackage[small]{caption}
\usepackage{graphicx}
\usepackage{amsmath}
\usepackage{amsthm}
\usepackage{booktabs}
\usepackage{algorithm}
\usepackage{algorithmic}
\usepackage[switch]{lineno}
\usepackage{natbib} 
\usepackage{amssymb,nccmath}

\usepackage{amsmath,amsfonts,bm}

\def\eqref#1{equation~\ref{#1}}

\def\1{\bm{1}}

\def\Ab{{\bm{A}}}
\def\Bb{{\bm{B}}}
\def\Cb{{\bm{C}}}
\def\Db{{\bm{D}}}

\def\Gb{{\bm{G}}}

\def\Ib{{\bm{I}}}

\def\Kb{{\bm{K}}}
\def\Lb{{\bm{L}}}
\def\Mb{{\bm{M}}}

\def\Ob{{\bm{O}}}
\def\Pb{{\bm{P}}}
\def\Qb{{\bm{Q}}}

\def\Vb{{\bm{V}}}
\def\Wb{{\bm{W}}}
\def\Xb{{\bm{X}}}
\def\Yb{{\bm{Y}}}
\def\Zb{{\bm{Z}}}

\DeclareMathAlphabet{\mathsfit}{\encodingdefault}{\sfdefault}{m}{sl}
\SetMathAlphabet{\mathsfit}{bold}{\encodingdefault}{\sfdefault}{bx}{n}

\def\RR{{\mathbb{R}}}

\DeclareMathOperator*{\argmin}{arg\,min}

\usepackage{pifont}
\usepackage{wrapfig,lipsum}       
\usepackage{subcaption}
\usepackage[symbol]{footmisc}

\newcommand{\xmark}{\ding{55}}%

\title{LongVQ: \underline{Long} Sequence Modeling with \underline{V}ector \underline{Q}uantization \\ on Structured Memory}

\author{
Zicheng Liu$^{1,2}$\and
Li Wang$^2$\and 
Siyuan Li$^{1,2}$\and
Zedong Wang$^2$\and
Haitao Lin$^{1,2}$\And
\stepcounter{footnote}Stan Z. Li$^{2}$\thanks{Corrsponding author.}\\
\affiliations
$^1$Zhejiang University, College of Information Science and Electric Engineering, Hangzhou, China\\
$^2$AI Lab, Research Center for Industries of the Future, Westlake University, Hangzhou, China\\
\emails
\{liuzicheng, wangli, lisiyuan, wangzedong, linhaitao, stan.zq.li\}@westlake.edu.cn
}

\begin{document}

\maketitle

\begin{abstract}
Transformer models have been successful in various sequence processing tasks, but the self-attention mechanism's computational cost limits its practicality for long sequences. Although there are existing attention variants that improve computational efficiency, they have a limited ability to abstract global information effectively based on their hand-crafted mixing strategies. On the other hand, state-space models (SSMs) are tailored for long sequences but cannot capture complicated local information. Therefore, the combination of them as a unified token mixer is a trend in recent long-sequence models. However, the linearized attention degrades performance significantly even when equipped with SSMs.
To address the issue, we propose a new method called LongVQ. LongVQ uses the vector quantization (VQ) technique to compress the global abstraction as a length-fixed codebook, enabling the linear-time computation of the attention matrix. This technique effectively maintains dynamic global and local patterns, which helps to complement the lack of long-range dependency issues.
Our experiments on the Long Range Arena benchmark, autoregressive language modeling, and image and speech classification demonstrate the effectiveness of LongVQ. Our model achieves significant improvements over other sequence models, including variants of Transformers, Convolutions, and recent State Space Models.
\end{abstract}

\section{Introduction}
\label{sec.intro}
Researchers and practitioners have been focusing on long-range dependency learning in sequence modeling. To tackle this challenge, popular sequence models like LSTM~\citep{hochreiter1997long} and GRU~\citep{chung2014empirical} have been introduced. These models can capture the relationships among input tokens (such as words~\citep{ainslie2020etc}, pixels~\citep{dosovitskiy2021vit}, and speech signals~\citep{pascual2019speech}) during the learning process. By utilizing contextual information, these models can learn intricate patterns of token dependencies in training samples. However, traditional sequential models are inefficient due to their non-parallelizability and limited memory. Consequently, scaling up the model parameters fails to enhance performance and significantly slows down training and inference speed. In response to the limitations of traditional sequence models, the Transformer~\citep{vaswani2017attention} architecture was introduced to achieve faster training and SotA performance. A key innovation enabling their parallelizability and superiority is the self-attention mechanism with positional encoding, which explicitly learns dependencies between all combinations of input tokens with order information encoded, aiming to create flexible kernels for sequences. 

\begin{figure*}
    \centering
    \includegraphics[width=1.0\linewidth]{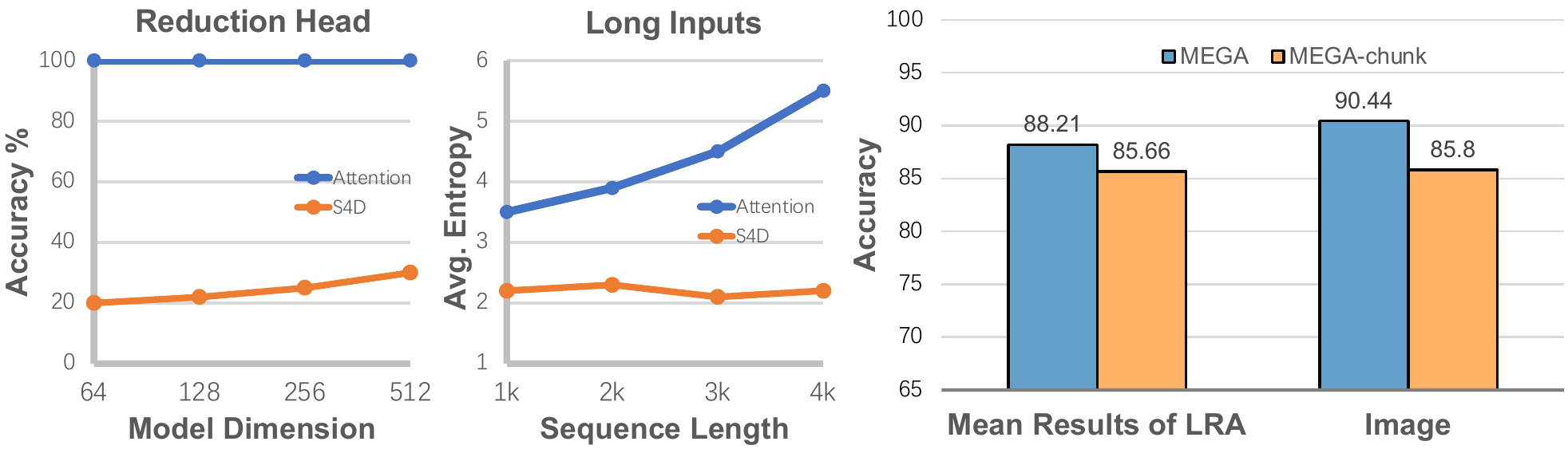}
    \vspace{-1.5em}
    \caption{Illustration of the complementary property of Attenion and State Space Model. Taking Reduction Head~\citep{fu2023h3} and Long Inputs as tasks on Text data: self-attention has a decisive ability to sift through information, but gets lost in long sequences, \textit{i.e.}, high entropy in attention matrix; SSM vice versa (\textit{left} and \textit{middle}). Generally, linear attention causes large performance degradation for the SSM-attention hybrid model (\textit{right}). The first two experiments are conducted with a two-layer model with 128 embedding dimensions.}
    \label{fig:intro}
\end{figure*}

Transformers have revolutionized natural language processing (NLP) and achieved significant breakthroughs. However, they have a major drawback when dealing with very long input sequences that may contain millions of tokens. That is a quadratic $\mathcal{O}(\Lb^2)$ complexity relative to the input sequence length in the self-attention mechanism, resulting in rapid scaling of compute and memory needs with sequence length $\Lb$.
Thus, many researchers in recent years focused on improving Transformer efficiency. Proposed techniques include approximating the costly self-attention matrix through various means like sparsity~\citep{tay2020sparse,liu2022automix} or conditional probability assumptions~\citep{ren2021combiner}. Other innovations like matrix factorization, optimized memory access patterns, and architectural constraints have also been explored~\citep{wang2020linformer, kitaev2020reformer,peng2021random,beltagy2020longformer,zhang2021poolingformer,lingle2023transformervq}. While all of these operations reduce the complexity of the model, at the level of expressiveness, transformers can easily get lost in long sequence data. The \textit{middle} of Fig.~\ref{fig:intro} demonstrates attention will fail in the long sequence scenarios, \textit{e.g.,} resulting high-entropy attention matrix. In parallel, wholly novel models have been introduced to better handle long sequences. Models like applying linear recurrent units, which integrate state space models(SSMs) and polynomial projections, to reconstruct latent histories and demonstrate promising performance, especially on long-sequence tasks~\cite{gu2020hippo,gu2020improving,gu2021combining,gu2021efficiently,gu2022parameterization,fu2023h3,fu2023longconv,poli2023hyena}. As shown in Fig.~\ref{fig:intro}(\textit{left}), it has to be admitted, however, that the SSM model is far less capable of screening and selecting information than models with attention mechanisms~\citep{fu2023h3}.

Therefore, in order to combine the respective advantages of the SSMs and Transformers, some hybrid models have been proposed recently and also achieved good results on long sequence tasks~\citep{ma2022mega,zuo2022spade}. To accommodate the computational complexity of the SSM model, linear attention is generally used as a collocation. Nevertheless, we can clearly see in the experimental results that the overall performance has a noticeable degradation after switching self-attention to linear attention (see \textit{right} of Fig.~\ref{fig:intro}). The question that needs to be addressed is: \textbf{is it possible to combine the advantages of both and match the performance of a hybrid model with self-attention to linear complexity?}

To answer this question, we present \textit{\textbf{Long} Sequence Modeling with \textbf{V}ector \textbf{Q}uantization} (LongVQ), an attention-SSM hybrid model in linear time with respect to the sequence length. Our method combines vector quantized keys, localized positional biases, and a dynamic fixed-size memory mechanism to store and retrieve complex data structures in long sequences efficiently. It has transformed how we approach data analysis and processing and unlocked new possibilities for various fields, including natural language processing, speech recognition, and image recognition. We benchmarked LongVQ on five popular datasets, including Long Range Arena (LRA)~\citep{tay2020lra}, which has diverse and challenging data modalities, the image dataset sCIFAR~\citep{cifar10}, the natural language datasets WikiText-103~\citep{merity2016wiki103} and enwik8, and the speech data SpeechCommand~\cite{warden2018speech}. In these datasets our method demonstrates promising performance.

\section{Related Work}
\subsection{Efficient Transformers}
\paragraph{Kernel method and clustering} 
Kernelizable attention, a technique introduced in studies by~\citep{katharopoulos2020transformers}, involves computing query and key features and then applying the same nonlinearity to both of them separately. It omits additional nonlinearities when computing attention weights. Clustering attention, as described by~\citep{vyas2020fast}, involves using VQ queries and can be kernelized.

\paragraph{Hierarchical Attention} 
Combiner, proposed by~\citep{ren2021combiner}, approximates softmax using a graphical model and enables decoder-only self-attention. H-Transformer-1D~\cite{zhu2021h} reduces the complexity of encoder-only self-attention. Transformer-LS~\cite{zhu2021long} downsamples long-range features in Transformers.

\paragraph{Compressed Attention} 
Compressive Transformers \citep{rae2019compressive,ainslie2020etc,zhu2021long}, learn a compression function for long-range features directly. On the other hand, LUNA~\citep{ma2021luna} uses cross-attention to compress long-range features into a recurrent state. Our LongVQ differs from compressive/recurrent Transformers in that it equates to quadratic-time attention over vector-quantized keys. This means that if the keys are already vector-quantized, the LongVQ cache reduces the cost to linear time without any loss. In contrast, Perceivers~\citep{jaegle2021perceiver} use cross-attention to attend to long sequences and compute self-attention over only a narrow stack of ``latents.'' However, LongVQ computes dense self-attention in linear time instead of just cross-attention. Therefore, while Perceivers' long-range layers incur a quadratic time complexity during sampling, LongVQ generates sequences in linear.

\subsection{State Space Models}
Recurrent neural networks and their linear counterparts, such as state-space models, can retain memory of the past. Among them, S4~\citep{gu2021efficiently} is notable because it can be implemented through convolutions thanks to its linear recurrence. However, the long convolution kernel for this model is as long as the input sequence, and its efficient computation requires sophisticated parameterization and approximation techniques. Although recent advances have found solutions to this issue, initializing these models still requires special effort \citep{gupta2022diagonal,gu2020improving}. Many of these models use the HiPPO~\cite{gu2020hippo} initialization mechanism, which aims to memorize historical data through projection to orthogonal polynomials. Based on a structure similar to SSM, an increasing number of works focusing on either linear recurrence or global convolution have been developed recently \citep{fu2023longconv,fu2023h3,poli2023hyena}.

\subsection{Vector Quantization}
Various techniques such as k-means, vector quantization, and codebooks have been used in Transformers for various applications aiming to learn semantic discrete tokens~\citep{roy2020efficient}. These techniques typically involve codebooks or similar structures within a Transformer architecture. Some models have proposed using a codebook outside the Transformer, such as when Transformers are priors for VQ-VAEs~\citep{oord2018neural,esser2021vqgan}. LongVQ utilizes one codebook within each layer and performs dense self-attention with SSMs in linear time, which distinguishes it from all of the aforementioned works.

\subsection{Data-dependent Gating}
Gated attention is a fusion of attention sublayers and GLU-based MLP sublayers, which was introduced in FLASH~\citep{hua2022flash}. Various gating mechanisms have been previously used to stabilize the training of Transformers and other sequence models. Examples of such models include GSS~\citet{mehta2022gss}, MEGA by~\citet{ma2022mega}, and RWKV by~\citet{peng2023rwkv}.

\section{Preliminary}
\label{sec.pre}
\subsection{Self-attention Mechanism}
\label{sec.attention}
\begin{figure*}
    \centering
    \includegraphics[width=0.85\linewidth]{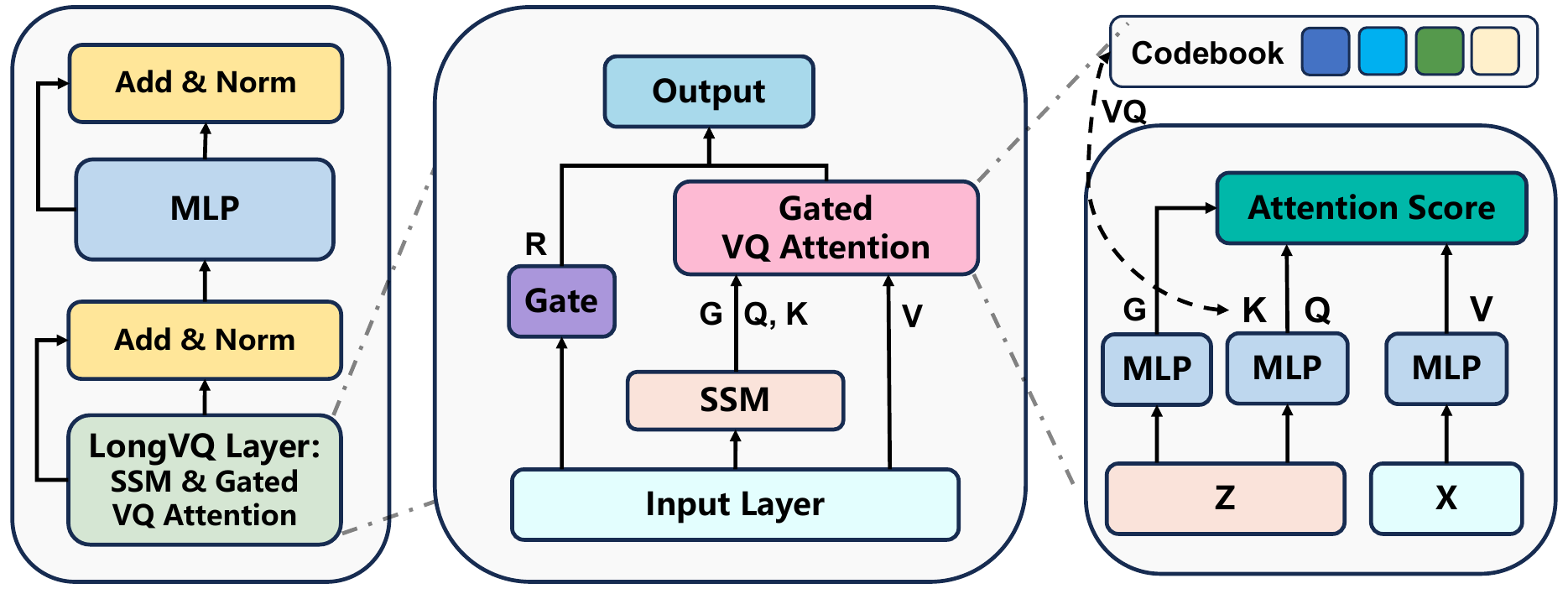}
    \caption{LongVQ – model architecture. Figure \textit{left} shows the overall architecture of each LongVQ block. Figure \textit{middle} illustrates the gated SSM-attention hybrid model with vector quantization method, while Figure \textit{right} displays the details of our proposed VQ-equipped Attention.}
    \label{fig:arch}
\end{figure*}

Given the input to the layer is $\Xb \in \RR^{L \times d}$, where $L$ is the sequence length and $d$ is the embedding dimension, then the self-attention mechanism outputs pair-wise score $\Mb$:
\begin{align} \label{eq:attention}
    &\Mb=\mathrm{Attn}(\Qb,\Kb,\Vb) = \sigma\left( \frac{\Qb \Kb^\top}{\sqrt{d}} \right) \Vb, \\
    &\text{where } \Qb = \Xb \Wb_q, \ \Kb = \Xb \Wb_k, \ \Vb = \Xb \Wb_v. \notag
\end{align}
Here $\Wb_q, \Wb_k, \Wb_v \in \RR^{d\times d}$ are learnable matrices, and $\sigma$ is the self-attention function \textit{e.g.,} softmax. The self-attention mechanism facilitates the computation of pair-wise relationships of all input tokens as cache, thereby enabling the modeling of long-range dependencies that surpasses the capabilities of recurrent neural networks. For example, denote the attention score $\Mb = \mathrm{softmax}(\Qb \Kb^\top / \sqrt{d}) \in \RR^{L \times L}$. Then, $\Mb_{ij}$ computes the similarity between the $i$-th and the $j$-th tokens.

\subsection{Linear Recurrent Models}

\textbf{Continuous linear state-space model.}
The center of the LSSL layer is a linear continuous-time SSM which maps a 1-dimensional function \(x(t)\) to \(y(t)\) through the hidden state $h(t) \in \mathbb{R}^{N \times 1}$. Concretely, LSSL can be formally defined as:
\begin{equation}
\label{eq:ssm}
\begin{aligned}
\hat{h}(t) &= \Ab h(t) + \Bb x(t) \\
y(t) &= \Cb h(t) + \Db x(t). \notag
\end{aligned}
\end{equation}
where \({\Ab} \in \mathbb{R}^{N\times N}, {\Bb}, {\Cb} \in \mathbb{R}^{N\times 1}, {\Db} \in \mathbb{R}^{1\times 1}\). The SSM can also be viewed as a convolution over the input signal.

Many researchers use Eq.~\ref{eq:ssm} to model long sequences. For instance, \citet{gu2020hippo} argue that randomly initialized parameters $\Ab$, $\Bb$, and $\Cb$ aren't effective in modeling long-range dependencies. To address this issue, they propose a set of matrices called HiPPO (high-order polynomial projection operators) for initializing $\Ab$. The HiPPO matrices are designed in such a way that the state $x(t)$ at time $t$ can remember the history of the input $u(t)$ up to time $t$.

\paragraph{Discrete-time form.}
The continuous-time LSSL in \ref{eq:ssm} needs to be discretized for real-world data. Using the bilinear method and a step size of $\Delta$:
\begin{align}
\label{eq:ssm-discrete}
    &x_k = \overline{\Ab} x_{k-1} + \overline{\Bb} u_{k}, \quad y_k = \overline{\Cb} x_k, \\
    &\text{where} ~ \overline{\Ab} = ({\Ib} - \Delta/2 \cdot {\Ab})^{-1} (\Ib + {\Delta}/2 \cdot {\Ab}), \notag \\ 
    &\quad\quad\ \ \ \overline{\Bb} = ({\Ib} - \Delta/2 \cdot {\Ab})^{-1} {\Delta} {\Bb}, \quad \overline{\Cb} = {\Cb}. \notag
\end{align}
We unroll the above recurrent representation, then we have:
\begin{align*}
    y_k = \overline{\Cb} \overline{\Ab}^k \overline{\Bb} u_0 + \cdots + \overline{\Cb} \overline{\Ab} \overline{\Bb} u_{k-1} + \overline{\Cb} \overline{\Bb} u_k.
\end{align*}
\paragraph{Convolution form.} This can be written as a convolutional representation $y = \overline{\Kb} * u$, where the convolution kernel
\begin{align} \label{eq:ssm-conv}
    &\overline{\Kb} \in \RR^L = \left( \overline{\Cb} \overline{\Bb}, \overline{\Cb} \overline{\Ab} \overline{\Bb}, \cdots, \overline{\Cb} \overline{\Ab}^{L-1} \overline{\Bb} \right).
\end{align}
Here, the SSM transformation can be viewed as embedding mapping from input to output by using a long convolution kernel. ``$*$'' is the discrete convolution operator, $u$ represents the input, and $y$ represents the corresponding output.
In Eq.~\ref{eq:ssm-conv}, the output $y$ can be efficiently computed if the convolution kernel $\overline{K}$ is known. However, computing the kernel can be challenging, with most existing algorithms having $O(L^2)$ time and space complexity.

\paragraph{Structured State Space Sequence Model}
\label{sec.longconv}
\citet{gu2021efficiently} develop the Structured State Space Sequence (S4) model to efficiently compute Eq.~\ref{eq:ssm-conv}. Specifically, $\Db$ is the residual connection, $\Cb$ in Eq.~\ref{eq:ssm} is randomly initialized, and $\Ab$ and $\Bb$ are initialized as structured matrices as following:
\begin{align} \label{eq:s4-init}
    &{\Ab} = {\Ab}^{(d_s)} - \Pb \Pb^\top, \quad {\Bb}_i = (2i+1)^{\frac{1}{2}}, \\
    &\text{where } {\Pb}_i = \left( i+1/2 \right)^{1/2}, \notag \\
    &{\Ab}^{(d_s)}_{ij} = -
    \begin{cases}
        (i+\frac{1}{2})^{\frac{1}{2}} (j+\frac{1}{2})^{\frac{1}{2}}, & i > j, \\
        \frac{1}{2}, & i = j, \\
        -(i+\frac{1}{2})^{\frac{1}{2}} (j+\frac{1}{2})^{\frac{1}{2}}, & i < j. \\
    \end{cases} \notag
\end{align}
The convolution can be computed efficiently with Fast Fourier Transforms (FFTs) when the kernel $K$ is known. However, for sequence length \(L\), computing the kernel \(K\) takes \(\mathcal{O}(LN^2)\) operations for unstructured \(\bar{\Ab}\). The paper proposed an efficient albeit complex algorithm for a class of structured \(\bar{A}\) with \(\tilde{\mathcal{O}}(N+L)\).

\section{Sequence Modeling with Vector Quantization on Structured Memory}
This section describes our proposed model, \textit{long sequence modeling with vector quantization}~(LongVQ). We first introduce the linear-time VQ attention equipped with the SSM model. Then, we present the whole block of LongVQ, including feed-forward and normalization layers. At last, we provide the network blocks and training algorithm of LongVQ.

\subsection{Vector Quantization}
Vector quantization (VQ) is a widely used information compression technique \citep{lingle2023transformervq}. In our work, VQ will be employed in self-attention layers to ensure linear complexity and make SSM operators more expressive.

\paragraph{What can VQ bring to SSM-attention hybrid?}
As discussed in Sec.~\ref{sec.intro}, SSM and self-attention mechanisms each have their own limitations despite their wide applications and impressive successes in sequence modeling. The hybrid models enjoy both side benefits by leveraging their properties to complement each other. However, the linear attention, \textit{e.g.,} chunk-wise attention, degrades the performance dramatically~\citep{ma2022mega}. The linear mixing methods designed artificially may lose the global view. VQ is a two-way solution that retains linear complexity and dynamically compresses global information to address this issue. The resulting model benefits from strong inductive bias while also being able to efficiently learn complex dependency patterns.

\paragraph{Vector Quantizers and Training}
A \emph{vector quantizer} is a function $\text{VQ}(\cdot; \mathbf{C})$ compresses a continuous input vector $x$ to a discrete embedding space $\hat{x}\in\RR^{S\times D}$ with fixed size: 
\begin{align}
    z &= \argmin_{s} || \mathbf{x} - \mathbf{C}_{s} ||^{2}, \\
    \hat{\mathbf{x}} &= \mathbf{C}_{z},
\end{align}
where $\mathbf{C} \in \mathbb{R}^{S \times D}$ is known as the \emph{codebook}. The row indices $\{0, \ldots, S-1\}$ of $\mathbf{C}$ are called \emph{shortcodes}, and the rows themselves are called \emph{codewords}. 
%
Because of the non-differentiable nature of the quantization operation, the straight-through estimator is typically used to update the network and \emph{codebook} in an end-to-end manner:
\begin{align}
z &= \argmin_{s} || \mathbf{x} - \mathbf{C}_{s} ||^{2}, \\
\hat{\mathbf{x}} &= \mathbf{x} + \text{SG}(\mathbf{C}_{z} - \mathbf{x}),
\end{align}
where $\mathrm{SG}(\cdot)$ denotes for stop-gradient operator.Since the computation of attention varies with the length of the sequence, linear complexity can be achieved as long as we can compress the $\Kb$ to a constant length.

\subsection{LongVQ Layer}
The gated attention method in LongVQ is inspired by MEGA~\citep{ma2022mega} and Gated Attention Unit~\citep{hua2022transformer} as the basic block in architectures. With an SSM layer embedded to capture long-term patterns, the input of our Gated VQ attention $\Gb, \Qb, \Kb, \Vb$ can be computed as follows:
\begin{align}
    &\Zb = \phi_{\mathrm{silu}}(\mathrm{SSM}(\Xb)), \\
    &\Gb_a = \phi_{\mathrm{silu}}(\mathrm{Linear}(\Zb)), \\
    &\Qb = \mathrm{Linear}(\Zb), \\
    &\Kb = \mathrm{Linear}(\Zb), \\
    &\Vb = \phi_{\mathrm{silu}}(\mathrm{Linear}(\Xb)),
\end{align}
where $\Zb$ can be regarded as the structured memory compresses the whole sequence. $\Gb_a$ is the gating variable that empowers the expressiveness of attention. Note that $\phi_{silu}$ is the self-gated activation function (SiLU)~\citep{ramachandran2017searching}. The output of Gated Attention is:
\begin{equation}
    \Ob_a = \Gb_a \odot \mathrm{Attn}(\Qb,\Kb,\Vb),
\end{equation}
where $\odot$ denotes the element-wise production.
\paragraph{Linear-time VQ Attention}
Note that the above computation of attention is still in quadratic complexity. Recall the vector quantization process in Sec.\ref{sec.pre}. By following \citep{oord2018neural},we could compress the $\Kb$ into a dynamic codebook with a fixed length:
\begin{equation}
    \hat{\Kb} = \mathrm{STVQ}(\Kb;\Cb),
\end{equation}
where $\mathrm{STVQ}(\cdot;\Cb)$ represents the row-wise application of vector-quantization with a straight-through gradient estimator. Thus, the attention score can be refactored as:
\begin{align}
    \mathrm{Attn}(\Qb,\hat{\Kb},\Vb) &= \phi_{\mathrm{element}}(\Qb\hat{\Kb}^{\top})\Vb, \\
    &=\phi_{\mathrm{element}}(\Qb\hat{\Cb}^{\top})\Delta\Vb,
\end{align}
where $\Delta$ could be regarded as \textit{Kronecker delta function} that builds a mapping between each token in $\Kb$ and code in $\Cb$. Here, we take the element-wise activation function as an example, \textit{i.e.,} $\phi_{\mathrm{element}}=\mathrm{relu}^2(\cdot)$. The resulting attention computation achieves linear-time complexity for efficient sequence modeling. 
\paragraph{Local window} If $\Cb\ll\Lb$, the code will appear uniformly throughout the sequence and will also cause the model to lose the ability to distinguish between near and far. To address this, a local bias $\Bb$ is added to the VQ attention for enhancing local information:
\begin{equation}
    \mathrm{Attn}(\Qb,\hat{\Kb},\Vb,\Bb)=\phi_{\mathrm{element}}(\Qb\hat{\Kb}^{\top}+\Bb)\Vb.
\end{equation}
Moreover, we add an output gate $\Gb_o$ after the attention to control the final output $\Ob$ dependent on the input $\Xb$:
\begin{align}
    \Gb_o &= \phi_{\mathrm{sigmoid}}\mathrm{Linear}(X), \\
    \Ob &= \Gb_o \odot \Ob_a + (1-\Gb_o) \odot \Xb.
    \label{eq:residual}
\end{align}
Fig~\ref{fig:arch} \textit{middle} plots the block design of LongVQ architecture.

\subsection{LongVQ Blocks}
The LongVQ layer can be used instead of regular self-attention in the Transformer model. Once added, it is followed by position-wise feed-forward networks (FFNs) and normalization layers to create a LongVQ block. As the gated residual connection is already included in equation \ref{eq:residual}, the original residual connection is omitted, and a normalization layer is directly applied to $\Yb$:
\begin{align}
    \Yb &= \mathrm{Norm}(\mathrm{LongVQ}(\Xb)), \\
    \Yb' &= \mathrm{Norm}(\mathrm{FFN}(\Yb)),
\end{align}
where $\Yb'$ is the output of a single LongVQ block. The overall LongVQ block is shown in Fig~\ref{fig:arch} \textit{left}. 
\paragraph{Training loss}
Let $\theta$ denote the set of non-codebook parameters, and $C$ denote the set of layers' codebooks. For modeling sequences $\Xb$ for classification, our loss can be defined:
\begin{equation}
    \mathcal{\Lb}_{\mathrm{LongVQ}} = \mathcal{\Lb}_{\mathrm{CE}}(\Xb; \theta, \Cb) + \gamma \mathcal{\Lb}_{\mathrm{VQ}}(\Xb; \theta, \Cb),
\end{equation}
where $\gamma$ is a hyperparameter known as the commit loss coefficient. The $\mathcal{\Lb}_{\mathrm{CE}}$ is the classification loss, and the $\mathcal{\Lb}_{\mathrm{VQ}}$ is employ mean squared error (MSE) between $\Kb$ and $\Cb$:
\begin{equation}
    \mathcal{\Lb}_{\mathrm{VQ}} = \mathrm{MSE}(\Kb, \mathrm{SG}(\Cb)).
\end{equation}
The main cross-entropy loss is augmented with commitment losses to update the codebook across all layers. As \citet{oord2018neural}, codebooks are parameterized through smoothed quantizer statistics.

\section{Experiments}
To evaluate LongVQ, we conduct experiments on five benchmark sequence modeling tasks across various data types, comparing them with current state-of-the-art models on each task. In all cases, we use codebook size 512. All experiments were realized based on NVIDIA A100-80G and Pytorch. We used float32 parameters, with bfloat16 precision for most computations. We adopt AdamW as the optimizer with a gradient clip of 0.1. The codebook commit coefficient was always $\gamma=0.0001$, and the codebook EMA rate was always $\eta=0.99$. All models were trained with a global batch size of 128 sequences. In our table, bold indicates the best-performing model and underlines the second-best.

\section{Pixel-level sequential image classification}
Let's begin with tasks related to image classification, where images are treated as a sequence of pixels in one dimension. In such tasks, models are not allowed to use any two-dimensional bias from the image. Therefore, the model should be capable of identifying patterns at various timescales, including pixels that are close to the original image but far from each other in its sequence representation.
We assess our model's performance on the Sequential CIFAR-10 dataset, which is widely used as a standard benchmark for modeling long-term dependencies in RNNs. The CIFAR-10 dataset's standard train and test split is used, and 10\% of the training set is withheld as the validation set. To accomplish the classification task, we compute the mean of all tokens in the output sequences and feed the result into a fully connected layer to generate class logits. The test accuracy is reported based on the model with the highest validation accuracy.
\begin{table}[t]
    \centering
    \caption{Performance of pixel-level sequential classification on the sCIFAR dataset. Results of compared methods are taken from either the citation or \citet{hasani2022liquid}.}
    \vspace{-0.75em}
    \label{tab:scifar}
        \vskip 0.1in
        \begin{small}
            \begin{tabular}{lccc}
                \toprule
                \textbf{Model} & \textbf{Accuracy} (\%) \\
                \midrule
                \emph{Attention}: & \\
                Transformer \citep{trinh2018learning} & 62.20 \\
                \midrule
                \emph{RNN}: & \\
                LSTM \citep{hochreiter1997long} & 63.01 \\
                r-LSTM \citep{trinh2018learning} & 72.20 \\
                UR-GRU \citep{gu2020improving} & 74.40 \\
                HiPPO-RNN \citep{gu2020hippo} & 61.10 \\
                LipschitzRNN \citep{erichson2020lipschitz} & 64.20 \\
                \midrule
                \emph{State Space Models}: & \\
                S4 \citep{gu2022parameterization} & 91.80 \\
                S4D \citep{gu2022parameterization} & 90.69 \\
                S5 \citep{smith2022simplified} &  90.10 \\
                Liquid-S4 \citep{hasani2022liquid} & \underline{92.02} \\
                \midrule
                \emph{Convolution}: & \\
                TrellisNet \citep{bai2018trellis} & 73.42 \\
                CKConv \citep{li2022makes} & 63.74 \\
                FlexConv \citep{romero2021flexconv} & 80.82 \\
                \midrule
                \emph{Hybrid}: & \\
                \textbf{LongVQ (Ours)} & \textbf{92.55} \\
                \bottomrule
            \end{tabular}
        \end{small} 
\end{table}
\begin{table*}[t] 
    \centering
    \vspace{-0.5em}
    \caption{Performance of predicting outcomes of list operations in the LRA~\citep{tay2020lra} benchmark. \textbf{Bold} indicates the best-performing model and \ul{underlines} the second best. Results are taken from either the citation. The training speed and peak memory consumption comparison on the Text task with the input length of 4K.}
    \vspace{-0.5em}
     \label{tab:lra}
    \begin{small}
        \begin{tabular}{@{} l | c c c c c c | c c c @{}}
        \toprule
        \textbf{Models}                                     & \textbf{ListOps}                   & \textbf{Text}  & \textbf{Retrieval} & \textbf{Image} & \textbf{Pathfinder} & \textbf{PathX} & \textbf{Avg.}  & \textbf{Speed}       & \textbf{Mem.}        \\ \hline
        \emph{Attention}:                                   &                                    &       &           &       &            &       &       &             &             \\
        Transformer \citep{vaswani2017attention}            & 36.37                              & 64.27 & 57.46     & 42.44 & 71.40      & \xmark     & 54.39 & 1.0$\times$ & 1.0$\times$ \\
        Local Attention \citep{tay2020lra}                 & 15.82                              & 63.98 & 52.98     & 41.46 & 66.63      & \xmark     & 46.06 & 5.3$\times$ & 0.1$\times$ \\
        Linear Trans. \citep{katharopoulos2020transformers} & 16.13                              & 65.90 & 53.09     & 42.34 & 75.30      & \xmark     & 50.55 & 4.7$\times$ & 0.1$\times$ \\
        Linformer \citep{wang2020linformer}                 & 35.70                              & 53.94 & 52.27     & 38.56 & 76.34      & \xmark     & 51.36 & 5.5$\times$ & 0.1$\times$ \\
        Sparse Transformer \citep{child1904generating}      & 17.07                              & 63.58 & 59.59     & 44.24 & 71.71      & \xmark     & 51.24 & 4.2$\times$ & 0.2$\times$ \\
        Performer  \citep{choromanski2020rethinking}        & 18.01                              & 65.40 & 53.82     & 42.77 & 77.05      & \xmark     & 51.41 & 5.7$\times$ & 0.1$\times$ \\
        Sinkhorn Transformer \citep{tay2020sparse}          & 33.67                              & 61.20 & 53.83     & 41.23 & 67.45      & \xmark     & 51.39 & 3.8$\times$ & 0.1$\times$ \\
        Longformer \citep{beltagy2020longformer}            & 35.63                              & 64.02 & 59.29     & 40.83 & 74.87      & \xmark     & 55.01 & 1.1$\times$ & 0.3$\times$ \\
        BigBird \citep{zaheer2020big}                       & 36.05                              & 64.02 & 59.29     & 40.83 & 74.87      & \xmark     & 55.01 & 1.1$\times$ & 0.2$\times$ \\
        Luna-256 \citep{ma2021luna}                          & 37.25                              & 65.78 & 79.56     & 47.86 & 78.55      & \xmark     & 61.95 & 4.9$\times$ & 0.1$\times$ \\
        Reformer \citep{kitaev2020reformer}                 & 37.27                              & 56.10 & 53.40     & 38.07 & 68.50      & \xmark     & 50.67 & 0.8$\times$ & 0.3$\times$ \\ \hline
        \emph{State Space Models}:                          &                                    &       &           &       &            &       &       &             &             \\
        S4 \citep{gu2022parameterization}                   & 59.60                              & 86.82 & 76.02     & 87.09 & 87.26      & 86.26 & 80.50 & -           & -           \\
        DSS \citep{gupta2022diagonal}                       & 57.60                               & 59.60 & 86.82     & \underline{90.90} & 88.65      & 94.20 & 86.09 & 4.8$\times$ & 0.1$\times$ \\
        S4D \citep{gu2022parameterization}                  & \underline{60.18}                              & 87.34 & 91.09     & 87.83 & 93.78      & 92.80 & 85.50 & -           & -           \\ \hline
        \emph{Hybrid}:                                 &                                    &       &           &       &            &       &       &             &             \\
        Mega-chunk \citep{ma2022mega}                & 58.76                              & \textbf{90.19} & \underline{91.25}     & 85.80 & 94.41      & 93.81 & 85.66 & 5.5$\times$ & 0.1$\times$ \\
        SPADE \citep{zuo2022spade}                         & 59.70                              & 87.55 & 90.13     & 89.11 & \underline{96.42}      & \underline{94.22} & \underline{86.19} & 5.2$\times$ & 0.1$\times$ \\
        \textbf{LongVQ (Ours)}                             & \textbf{61.02}    & \underline{89.72} & \textbf{91.45}     & \textbf{91.35} & \textbf{96.70}      & \textbf{95.85} & \textbf{87.68} & 5.3$\times$ & 0.1$\times$ \\ \bottomrule
        \end{tabular}
    \end{small}
\end{table*}

\paragraph{Results} 
The Table~\ref{tab:scifar} displays the results. Our model has achieved state-of-the-art performance and the best test accuracy on the sequence classification task, surpassing numerous strong competitors such as Transformers~\citep{vaswani2017attention}, RNNs, state space models, and other convolutional models. Specifically, the LongVQ model has outperformed the previous convolution-based models by more than ten percentage points. 
It is noteworthy to mention that our model has achieved impressive results by surpassing the previously established performance benchmark, despite utilizing a relatively simple architecture. The model primarily uses a hybrid approach that compresses long historical information based on the output of SSM. Our most effective model comprises of ten LongVQ blocks, which play a significant role in achieving exceptional performance.

\subsection{Long-Context Sequence Modeling}
We have started conducting experiments by evaluating sequence models on the Long Range Arena (LRA) benchmark, which was recently introduced by~\citet{tay2020lra}. The benchmark aims to evaluate the performance of sequence models in long-context scenarios. The benchmark consists of a total of six tasks, each of which is designed to test different aspects of long-sequence modeling. These tasks include ListOps, Text, Retrieval, Image, Pathfinder, and Path-X. To ensure a comprehensive evaluation of our model, the tasks involve input sequences of varying lengths, ranging from 1K to 16K tokens. Additionally, the tasks cover a diverse range of data types and modalities, meaning that models must be capable of handling text, images, and other forms of input in order to perform well on the benchmark.

\paragraph{Results} In Table~\ref{tab:lra}, LongVQ is compared to various baselines, such as Transformer and its efficient versions, as well as the top-performing S4 models. Specifically, LongVQ outperforms the second-ranked model by 1.84\%, 0.79\%, 0.20\%, 0.45\%, 0.28\%, 3.63\% on each of the six datasets.
To ensure a just comparison, we equalize the number of layers and model dimensions on each task so that LongVQ has a similar number of parameters with S4. The results are an average of five runs with distinct random seeds, and the tuning information and model details are given in the Appendix.
The LongVQ has shown remarkable performance over all six tasks, with an average performance of 88.36\%, outperforming all the other baselines. Furthermore, we have evaluated the speed and memory efficiency of LongVQ using the byte-level classification task with an input length of 4K. Our VQ mechanism has proven to be highly efficient, being 5.3 times faster and consuming only 10\% as much memory as the vanilla Transformer. It is worth noting that the LongVQ layer, with its VQ-equipped hybrid design, is even more efficient than both Transformers and State Space Models by a large margin.

\begin{table}[b]
    \centering
    \caption{Performance and training speed on WikiText-103 dataset.}
    \vspace{-0.5em}
\label{tab:wiki103}
    \begin{tabular}{l|ccc}
    \toprule
                   & \multicolumn{3}{c}{\textbf{WikiText-103}} \\
    \textbf{Model}          & \textbf{\#Param.}         & \textbf{PPL}         & \textbf{Speed}  \\ \hline
    Transformer-adaptive  &247M       &18.66        &5.6 k t/s \\
    Transformer-XL & 257M             & 18.30       & - \\
    S4\citep{gu2020improving}    & 249M             & 20.95      & -   \\
    Mega-chunk\citep{ma2022mega} & 252M             & \underline{18.07}     & 48k t/s    \\
    \textbf{LongVQ (Ours)}       & 254M             & \textbf{17.85}       & 50k t/s  \\ \bottomrule
    \end{tabular}
\end{table}

\section{Auto-regressive Language Modeling}
By following~\citet{ma2022mega,lingle2023transformervq}, we evaluate LongVQ on two established language modeling benchmarks, \textit{i.e.}, WikiText-103~\citep{merity2016wiki103} and enwik8~\citep{2011largetext}, which are next-token prediction tasks. WikiText-103 is a word-level language modeling dataset containing 103M training tokens from Wikipedia articles. Following previous work~\citep{baevski2019adaptive}, our approach involves the utilization of adaptive softmax and input embeddings, and we work with a vocabulary of 260K tokens. enwik8, on the other hand, is a challenging benchmark for character-level language modeling. It comprises 100M tokens of unprocessed Wikipedia articles and has a vocabulary size of around 200. We divide the test data into segments and sequentially handle each segment during testing. 
\paragraph{Results} In Table~\ref{tab:wiki103}, we compare with previous top-performing models that are designed to take advantage of longer context, including Transformers~\citep{baevski2019adaptive}, Transformer-XL and S4~\citep{gu2021efficiently}. The LongVQ model has shown impressive performance on both WikiText-103 and enwik8 datasets, surpassing the baseline models by a considerable margin. Additionally, our model is able to achieve a faster (almost 10$\times$) inference speed compared to the Pure Transformer model. One of the key benefits of the hybrid design of the SSM layer and VQ technique is that it enables LongVQ to naturally handle length extrapolation at inference time, allowing it to process longer sequences than those seen during training.

\begin{table}[ht]
    \centering
    \caption{Testing bits-per-byte on Enwik8 dataset.}
    \vspace{-0.5em}
\label{tab:enwiki}
    \begin{tabular}{l|cc}
    \toprule
                   & \multicolumn{2}{c}{\textbf{enwik8}} \\
    \textbf{Model}          & \textbf{\#Param.}       & \textbf{PPL}       \\ \hline
    Transformer-XL & 41M           & 1.06      \\
    Mega~\citep{ma2022mega}           & 39M           & 1.02      \\
    Transformer-VQ~\citep{lingle2023transformervq} & 190M           & \underline{0.99}      \\
    \textbf{LongVQ (Ours)}         & 47M           & \textbf{0.97}      \\ \bottomrule
    \end{tabular}
    \vspace{-0.5em}
\end{table}
\section{Raw Speech Classification}
We aim to assess LongVQ's ability to model speech signals over a long range by utilizing it to classify unprocessed speech signals that have a length of 16000 instead of relying on conventional preprocessing methods, such as converting them to MFCC features. As per \citet{gu2021efficiently} approach, we conduct speech classification on the Speech Commands dataset's SC10 subset, which was introduced by \citet{warden2018speech}. As reported in \citep{ma2022mega}, the Mega-chunk uses a chunk size of 1000 to enable processing the data. 
\paragraph{Results} As shown in Table~\ref{tab:sc}, our LongVQ model with 542K parameters can achieve an accuracy of 97.67, which is the state-of-the-art method in this table. This is mainly because the large number of continuous and low-frequency signals in speech are well-suited to be processed using VQ. Coupled with SSM's ability to extract global information, it is only natural that such results are achieved.

\begin{table}[ht]
    \centering
    \caption{Accuracy on Speech Commands dataset.}
    \vspace{-0.5em}
\label{tab:sc}
    \begin{tabular}{l|cc}
    \toprule
                & \multicolumn{2}{c}{\textbf{SpeechCommand-Raw}} \\
    \textbf{Model}       & \textbf{\#Param.}          & \textbf{Accuracy}          \\ \hline
    Transformer & 786K              &  \xmark                 \\
    S4~\citep{gu2021efficiently}     & 300K              & \underline{97.50}             \\
    Mega~\citep{ma2022mega}  & -              & \xmark             \\
    Mega-chunk~\citep{ma2022mega}  & 476K              & 96.03             \\
    \textbf{LongVQ (Ours)}     & 542K              & \textbf{97.67}             \\ \bottomrule
    \end{tabular}
    \vspace{-0.5em}
\end{table}

\section{Ablation Study}
The most critical parameters in LongVQ are the codebook size and the commit loss hyperparameter $\gamma$. So, we focus our analysis on these two parts to answer two more important questions: \textbf{(1) Is it possible to set parameters for data types in the codebook? (2) What is the best attention function for LongVQ?}

\subsection{Codebook Size}
We take the subset of LRA to conduct the ablation study. As shown in Table~\ref{tab:codebook}, larger codebook sizes may allow more flexible attention patterns and improve the gradients' fidelity, and improve final performance. However, we can set the size of the codebook according to the resource requirements based on the data characteristics. For example, from images to logical reasoning, images with lower information density can use a lower size and vice versa.

\begin{table}[ht]
    \centering
    \caption{Ablation study on Codebook size and Attention functions.}
    \vspace{-0.5em}
\label{tab:codebook}
    \begin{tabular}{l|ccc|cc}
    \toprule
     & \multicolumn{3}{c|}{\textbf{Codebook Size}}          & \multicolumn{2}{c}{\textbf{Attention Function}} \\
    \textbf{Settings} & \textbf{256}   & \textbf{512}   &\textbf{1024}  & \textbf{softmax} & \textbf{laplace} \\ \hline
    Image    & 91.32 & 91.35 & 91.58 & 91.35   & 90.15   \\
    Text     & 86.26 & 89.72 & 89.75 & 89.72   & 88.27   \\
    ListOps  & 58.16 & 61.02 & 61.10 & 61.02   & 58.14   \\ \bottomrule
    \end{tabular}
    \vspace{-0.5em}
\end{table}
\vspace{-1em}
\subsection{Attention function}
We tested two attention functions, softmax and laplace, to evaluate performance. Based on LongVQ, our findings differ from Mega's~\citep{ma2022mega}. Due to its compressed codebook on $\Kb$, we found that LongVQ prefers softmax as the attention function for different data types; the results are shown in Table~\ref{tab:codebook}. This is mainly due to the existence of VQ technology, which dynamically compresses the data into codes, and softmax, which has the ability to filter information with strong sparsity, is more suitable for the data after compression. Hence, our LongVQ does not need parameter search in the choice of attention function.

\section{Conclusion and Limitations}
We presented LongVQ for robust and efficient memorization of long-term patterns in sequential data sources. LongVQ is an SSM-attention hybrid architecture that computes softmax-based dense self-attention in linear time with respect to sequence length. Its superior performance is enabled by considering the global view of the SSM module and a gating mechanism. The vector quantization method to compress keys is the crucial point to making LongVQ more efficient when handling long sequences. By using the local window design, LongVQ can still capture long-range patterns even with small codebook sizes. Our large-scale and diverse experiments demonstrate that LongVQ is an efficient and flexible long sequence model with excellent performance on image, text, logical reasoning, and speech data.

However, LongVQ also has some obvious limitations in empirical implementations. LongVQ has a relatively large training difficulty, and due to the online training codebook, we need to handle the gradient size of different modules carefully. Moreover, the exponential moving average (EMA) technique is essential for stabilizing codebook updates, but it also limits the update speed simultaneously. Whether a better-stabilized training model can be used on codebooks is still a question worth considering. One last thing to consider is whether there is a better solution than a localized window that allows the model to distinguish between the same code in different locations. By embracing LongVQ as a starting point in the integration of VQ into hybrid models, we are taking the initial step towards achieving greater efficiency gains in the future. As we continue to work towards solving more issues and scaling up this model to larger sizes, we are paving the way for a brighter and more innovative tomorrow.





\newpage
\bibliographystyle{named}
\bibliography{ijcai24}
\newpage
\appendix
\onecolumn
\section{Experimental Details}
\subsection{Long Range Arena (LRA) and sCIFAR}
For all tasks, we closely follow \citet{tay2020lra} for details such as data preprocessing, data split, etc. The hyper-parameters of \textsc{LongVQ} models on these tasks are listed in Table~\ref{tab:lra:hyps}. The experimental configuration of sCIFAR follows the parameter settings of the image in LRA.

\begin{table}[!h]
\caption{Hyper-parameters of \textsc{LongVQ} models on LRA and raw speech classification tasks. BSZ is batch size, LR is learning rate and WD is weight decay.
BN, LN and SN refer to Batch Normalization, Layer Normalization and Scale Normalization.}
\label{tab:lra:hyps}
\centering
\resizebox{\columnwidth}{!}{
    \begin{tabular}{l|ccccccccccc}
    \toprule
    \textbf{Task}       & \textbf{Depth} & $d_\mathrm{model}$ & $d_\mathrm{FFN}$ & \textbf{Attn-FN} & \textbf{Norm} & \textbf{Pre-norm} & \textbf{BSZ} & \textbf{LR} & \textbf{Dropout} & \textbf{WD} & \textbf{Epochs} \\ \hline
    \textbf{ListOps}    & 6                               & 80                 & 160              & laplace                           & BN                             & False                              & 64                            & 0.001                        & 0.1                               & 0.01                         & 60                               \\
    \textbf{Text}       & 4                               & 128                & 256              & softmax                           & SN                             & False                              & 50                            & 0.004                        & 0.1                               & 0.01                         & 50                               \\
    \textbf{Retrieval}  & 6                               & 128                & 256              & softmax                           & SN                             & False                              & 64                            & 0.003                        & 0.1                               & 0.04                         & 40                               \\
    \textbf{Image}      & 8                               & 160                & 320              & laplace                           & BN                             & True                               & 50                            & 0.01                         & 0.0                               & 0.02                         & 200                              \\
    \textbf{Pathfinder} & 6                               & 128                & 256              & laplace                           & BN                             & True                               & 128                           & 0.01                         & 0.0                               & 0.01                         & 200                              \\
    \textbf{Path-X}     & 4                               & 64                 & 128              & laplace                           & BN                             & True                               & 128                           & 0.01                         & 0.0                               & 0.01                         & 100                              \\ \hline
    \textbf{SC}         & 6                               & 60                 & 120              & laplace                           & BN                             & True                               & 20                            & 0.01                         & 0.0                               & 0.01                         & 200   \\     \bottomrule                        
    \end{tabular}
}
\end{table}

\subsection{Language Modeling}
We use the data of WikiText-103 and enwik8 and their splits provided by \citet{ma2022mega}. At training time, we split the training data into segments; each segment contains $m$ consecutive chunks, where the chunk size is the effective attention length. Other training hyperparameters, including optimizer, learning rate scheduler, and architecture, are presented in Table~\ref{tab:lm:hyps}.

\begin{table}[!h]
\caption{Hyper-parameters of models for language modeling.}
\label{tab:lm:hyps}
\centering
    \begin{tabular}{l|ccc}
    \toprule
     & \textbf{WikiText-103}  & \textbf{enwik8} \\
    \midrule
    Batch Size $\times$ GPUs  & 6144 $\times$ 8 & 8192 $\times$ 8 \\
    Optimizer & AdamW & AdamW \\
    Learning Rate &  0.005 & 0.005 \\
    Adam-$\beta$ & $(0.9, 0.98)$ & $(0.9, 0.98)$ \\
    Learning Rate Decay & linear & linear \\
    Weight Decay & 0.1 & 0.1 \\
    Dropout & 0.3 & 0.1 \\
    Attention Dropout & 0.1 & 0.0 \\
    FFN Hidden Dropout & 0.1 & 0.0 \\
    Gradient Clipping & 1.0 & 1.0 \\
    Warmup steps & 24K & 24K \\
    Total updates & 400K & 400K \\
    \midrule
    Decoder Layers & 16 & 12 \\
    Model size & 1024 & 512 \\
    FFN Hidden size & 1536 & 1024 \\
    Shared Repr. size ($z$) & 256 & 128 \\
    Value Seq. size ($v$) & 1536 & 1024 \\
    EMA dimension ($h$) & 16 & 16 \\
    Codebook Size & 512 & 512 \\
    Total Parameters & 254M & 47M \\
    \bottomrule
    \end{tabular}
\end{table}

\end{document}